\documentclass[10pt]{article}
\usepackage[preprint]{tmlr}

\usepackage{amsmath,amsfonts,bm}

\def\eqref#1{equation~\ref{#1}}

\def\1{\bm{1}}

\DeclareMathAlphabet{\mathsfit}{\encodingdefault}{\sfdefault}{m}{sl}
\SetMathAlphabet{\mathsfit}{bold}{\encodingdefault}{\sfdefault}{bx}{n}

\usepackage{hyperref}
\usepackage{url}
\usepackage{graphicx}
\usepackage{microtype}
\usepackage{booktabs}
\usepackage{wrapfig}
\usepackage{multirow}
\usepackage{colortbl}
\usepackage{makecell}
\usepackage[export]{adjustbox}
\usepackage{arydshln}
\usepackage{subcaption}

\usepackage{amsmath,amsthm,amssymb}

\usepackage{tikz}
\usetikzlibrary{arrows.meta,positioning}

\theoremstyle{plain}
\newtheorem{theorem}{Theorem}
\newtheorem{proposition}{Proposition}

\theoremstyle{definition}
\newtheorem{definition}{Definition}

\title{The Self-Improvement Problem}

\author{Brian B. Moser$^{1,2}$,
Stanislav Frolov$^{1}$,
Tobias C. Nauen$^{1,2}$,
Federico Raue$^{1}$,
Andreas Dengel$^{1,2}$ \\
\\[-3pt]
$^{1}$\textbf{German Research Center for Artificial Intelligence (DFKI)} \\
$^{2}$\textbf{RPTU Kaiserslautern-Landau} \\
\\[-3pt]
\texttt{first.last@dfki.de}
}
\begin{document}

\maketitle

\begin{abstract}
With the rise of large language models, agentic systems are becoming increasingly autonomous, as they write code, call tools, revise prompts, and propose modifications to their own components.
This autonomy raises a fundamental question for safe self-improvement: can an automated gate like a reward model or an automated reviewer reliably decide whether a proposed update should be accepted?
The difficulty is structural.
The gate must be a terminating, computable procedure, while the property we ultimately care about, namely whether an update is a genuine semantic improvement, depends on potentially unbounded program behavior.
We formalize this mismatch by modeling agents as programs and defining improvement in terms of a semantic score that need not be computable. We study two evaluation abstractions.
The first is a certifier that approves or rejects an update.
The second is a search procedure that, given access to a proposal generator, attempts to output some improving candidate if one exists.
We show that, in the fully general setting, no terminating procedure can simultaneously guarantee reliability (never accepting a non-improvement) and non-conservatism (never missing an available improvement): self-referential constructions can force any such gate or searcher to fail.
Randomized gates with bounded error face the same limit.
We then identify a standard decidable regime used in practice (bounded evaluation and a finite proposal budget) under which exhaustive simulation yields a terminating search procedure with the desired guarantees.
We also show the limit of this recovery: a chain of accepted updates can raise the bounded score in every round while the true score falls.
Our results draw a boundary between unbounded, universal settings where correctness guarantees collapse and bounded settings where they can be recovered by design.
\end{abstract}

\section{Introduction}

Many modern ``agentic'' pipelines follow the same pattern: an agent proposes a self-update (a code patch, prompt rewrite, new checkpoint, tool policy, etc.), and an automated gate decides whether to apply it \citep{lu2024ai, yamada2025ai}.
The gate may be a reward model, a regression test suite, or an automated reviewer that must \emph{terminate} and output accept/reject.
The intended notion of improvement, however, is semantic: whether the updated agent is genuinely better depends on unbounded program behavior, long-horizon interactions, and open-ended tool use \citep{wang2023voyager}.
This mismatch creates a concrete failure mode that is now routine in practice:
\emph{once the proposal mechanism can anticipate or condition on the gate, it can ``optimize the gate'' rather than the intended objective.}
In other words, the proposer can exploit the evaluator as a predictable part of the environment.

This paper formalizes that failure mode and proves a sharp boundary: in a fully general setting, \emph{no terminating gate or search procedure can be both reliable and non-conservative.}
The core obstruction is not statistical; it is structural.
If candidates can depend (directly or indirectly) on the gate's decision, then self-reference yields diagonal constructions in which any total computable gate must fail, either by accepting a non-improvement or by rejecting (or missing) an available improvement.

We model an ``agent'' as an executable artifact (e.g., checkpoint + code) encoded by a program index $p\in\mathbb{N}$, with behavior given by the partial function $\varphi_p$.
A proposed update is another program index $q$.
The gate is a total computable procedure that must halt with a decision:
either a \emph{certifier} $V(p,q)\in\{0,1\}$ that approves/rejects a given candidate, or a \emph{searcher} $S(p,e)$ that, given access to a proposal enumerator $E_e$, attempts to output some improving proposal.
We measure ``true'' performance via a semantic score functional $J$, which may be uncomputable.

Our first theorem shows that even for an extremely simple semantic score $J$ (halting-and-output-$1$ on a fixed input),
there is no total computable certifier that accepts \emph{exactly} the strict improvements.
The proof builds a self-referential candidate whose behavior flips depending on whether it is accepted,
using Kleene's recursion theorem \citep{kleene1952introduction}.
This captures, in the starkest possible form, the modern evaluator-hacking phenomenon: the proposal mechanism can construct candidates that are ``good if rejected'' and ``bad if accepted.''

A natural response is to replace certification by search: rather than deciding each candidate, we ask for a total searcher that returns \emph{some} improvement whenever the proposal mechanism contains one.
Our second theorem shows that this does not help in full generality.
Even if the searcher is given the code of the proposal enumerator, self-reference allows the enumerator to propose a candidate whose behavior depends on the searcher's output, forcing any proposal-respecting total searcher to violate soundness or weak completeness.

However, these negative results rely on unbounded behavior and universal quantification over proposal mechanisms.
We then formalize a standard regime used in practice (bounded evaluation and a finite proposal budget) and show that the guarantees can be recovered \emph{relative to the bounded proxy}.
In that regime, exhaustive simulation yields a terminating searcher that is sound and weakly complete for the bounded score.
This draws a clean line between (i) an unbounded universal regime where correctness guarantees collapse, and (ii) a bounded regime where they can be recovered by design.

Two further results sharpen this picture.
First, randomness does not help.
A gate that flips coins and errs with probability at most $1/3$ can be converted into a deterministic sound and complete certifier, so it cannot exist either (Proposition~\ref{prop:randomized}).
Second, the bounded regime has a blind spot of its own.
We construct a chain of updates in which every step raises the bounded score and every step lowers the true score (Proposition~\ref{prop:drift}).
Guarantees relative to a proxy therefore do not control drift of the true objective across rounds.

\paragraph{Scope of the contribution.}
We want to be clear about what is new here and what is not.
The proof ingredients are classical: diagonalization, the $s$-$m$-$n$ theorem, and Kleene's recursion theorem \citep{kleene1952introduction}.
We do not claim new computability-theoretic techniques.
The contribution is the formalization.
It gives a minimal model of verification-gated self-improvement in which reliability and progress guarantees can be stated exactly.
It separates the impossibility in the unbounded setting from its recovery in the bounded setting.
It turns the folklore claim that evaluator hacking is unavoidable once the proposer can model the gate into a theorem with explicit assumptions.
The formalization also yields results beyond the single update step, namely the randomized-gate bound of Proposition~\ref{prop:randomized} and the drift construction of Proposition~\ref{prop:drift}.

In summary, we make three contributions:
\begin{itemize}
  \item \textbf{A minimal formalization of verification-gated self-improvement.}
  We model agent updates as program transformations, proposal generation as a computable enumerator, and gating as a terminating procedure that must decide whether to apply an update.

  \item \textbf{A sharp impossibility boundary driven by self-reference.}
  In the fully general (unbounded) setting, we prove that no total computable procedure can simultaneously guarantee
  \emph{reliability} (never accepting a non-improvement) and \emph{non-conservatism} (never rejecting or missing an available improvement),
  for either certification (Theorem~1) or proposal-respecting search (Theorem~2).
  The impossibility extends to randomized gates with bounded error (Proposition~\ref{prop:randomized}).

  \item \textbf{Decidability under standard restrictions.}
  Under bounded evaluation and a known finite proposal budget, exhaustive checking yields a terminating searcher that is sound and weakly complete with respect to the bounded score (Theorem~3),
  clarifying exactly which design constraints restore guarantees.
  We also show the limit of this recovery: repeated accepted updates can raise the bounded score while the true score falls (Proposition~\ref{prop:drift}).
\end{itemize}

A Lean~4 formalization of Theorems~1--3 accompanies this paper, available at \url{https://github.com/Brian-Moser/SelfImprovementProof}.

\begin{figure}[t]
    \centering
    \includegraphics[width=0.6\textwidth]{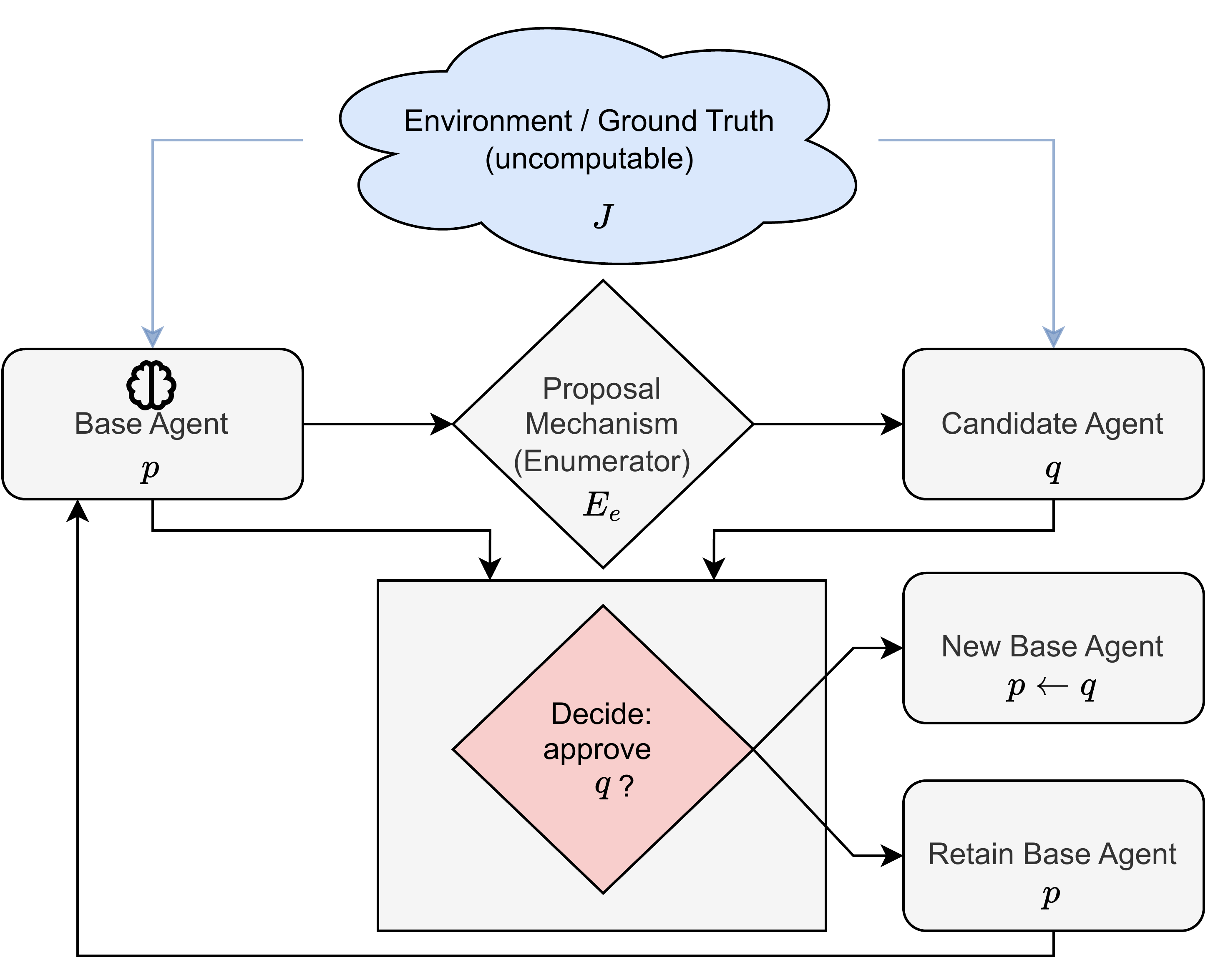}

    \caption{\textbf{The structural impossibility of verification-gated self-improvement.}
    The Proposal Mechanism (Enumerator $E_e$) generates a Candidate ($q$) whose behavior explicitly depends on the decision of the Gatekeeper (Searcher $S$).
    This self-referential loop forces any total computable gatekeeper to violate either soundness (by accepting a bad update) or completeness (by rejecting a valid improvement).}
    \label{fig:self_improvement_loop}
\end{figure}

\section{Preliminaries}

This section bridges the gap between common ML intuitions (models, checkpoints, loss functions) and the computability-theoretic formalization used in our impossibility result.
%
%
First, we model an ``agent'' as an executable artifact: e.g., a serialized model checkpoint plus inference code, a Python package, or a container image.
Any such finite binary string can be encoded as a natural number.
Consequently, we identify each agent with a program index $p \in \mathbb{N}$ (a G\"odel number \citep{godel1992formally}).

Let $\varphi_p$ denote the (partial) function computed by program $p$ under a fixed universal machine.
Given an input $x$ (e.g., a prompt, an environment state, or a task instance), $\varphi_p(x)$ denotes the output/action of the agent. Because an agent may run arbitrary code (loops, recursion, tool calls),
$\varphi_p(x)$ may \emph{diverge} (run forever). Thus $\varphi_p$ is in general a \emph{partial computable function}.
Before we approach the core contributions of this paper, let us introduce another fundamental tool \citep{kleene1952introduction}:

\paragraph{Kleene's Recursion Theorem (Fixed-Point Theorem).}
For any total computable function $f:\mathbb{N}\to\mathbb{N}$, there exists an index $e\in\mathbb{N}$ such that
\begin{equation}
\varphi_e \;=\; \varphi_{f(e)}.
\end{equation}
Intuitively, this guarantees the existence of self-referential programs whose behavior can depend on (an encoding of)
their own description. We use this theorem to construct adversarial candidates in the proofs of Theorems~1--2.


Another central theme in alignment and evaluation is the gap between what we \emph{wish} the agent to optimize, and  what we can \emph{compute} or \emph{verify} reliably.

\paragraph{Semantic objective (ground truth).}
We represent the intended notion of performance by a semantic score functional
\begin{equation}
J : \mathbb{N} \to \mathcal{S},
\end{equation}
where $\mathcal{S}$ is a score set, e.g., $\{0,1\}$ for success/failure, or $\mathbb{R}$ for reward.
Crucially, $J$ is allowed to be \emph{uncomputable} in general: perfect evaluation of real systems may require unbounded testing, reasoning about all inputs, or solving problems that are undecidable in the worst case.

\paragraph{Certifier (proxy).}
Since $J$ may be unavailable, the system may rely on a \emph{certifier}
\begin{equation}
V : \mathbb{N}\times \mathbb{N} \to \{0,1\},
\end{equation}
where $V(p,q)=1$ means ``approve updating agent $p$ to agent $q$.'' In ML practice, $V$ corresponds to
a reward model, a test suite, a static analyzer, or any automated gate that must return a decision.
Accordingly, we assume $V$ is \emph{total computable}: it always halts with an output.

\section{Self-Improvement Loop}

We view self-improvement abstractly as a process that proposes a candidate modification $q$ to replace
a current agent $p$, and applies the update only if $V(p,q)=1$.
To formalize the notion of ``improvement'' relative to the semantic objective $J$, we use:

\begin{definition}[Strict improvement]
Given $J:\mathbb{N}\to\mathcal{S}$ equipped with a strict order $>$, we say that $q$ is a
\emph{strict improvement} over $p$ if $J(q)>J(p)$.
\end{definition}

\noindent
The properties we would like a verification-based self-improvement system to satisfy are:

\begin{definition}[Soundness and completeness of a certifier]
Fix $J$ and strict improvement as above.
A total computable $V:\mathbb{N}\times\mathbb{N}\to\{0,1\}$ is
\begin{itemize}
  \item \emph{Sound} if $\forall p,q\ (V(p,q)=1 \Rightarrow J(q)>J(p))$.
  \item \emph{Complete} if $\forall p,q\ (J(q)>J(p) \Rightarrow V(p,q)=1)$.
\end{itemize}
\end{definition}

Soundness corresponds to ``no reward hacking / no false approvals'' (safety),
while completeness corresponds to ``do not block genuine progress'' (non-conservatism).
Our main theorem shows that, for sufficiently expressive agents (allowing self-reference),
these goals cannot be simultaneously achieved by any total computable certifier.

\begin{theorem}[No sound-and-complete improvement certifier]
Let programs be encoded by natural numbers $p \in \mathbb{N}$, and let $\varphi_p$ denote the (partial) function computed by program $p$.
Define the semantic score functional $J:\mathbb{N}\to\{0,1\}$ by
\begin{equation}
J(p) \;=\;
\begin{cases}
1 & \text{if $\varphi_p(0)$ halts and outputs $1$,}\\
0 & \text{otherwise.}
\end{cases}
\end{equation}
Call $q$ a \emph{strict improvement} over $p$ if $J(q) > J(p)$.
There is no total computable function $V:\mathbb{N}\times\mathbb{N}\to\{0,1\}$ such that for all $p,q$:
\begin{enumerate}
  \item \textbf{Soundness:} $V(p,q)=1 \implies J(q) > J(p)$,
  \item \textbf{Completeness:} $J(q) > J(p) \implies V(p,q)=1$.
\end{enumerate}
\end{theorem}

\begin{proof}
Assume for contradiction that such a total computable certifier $V$ exists.
Let $p$ be an index of a program that halts on input $0$ and outputs $0$.
Then, by definition of $J$, we have $J(p)=0$.
Informally, we want a program $q$ whose output on input $0$ flips depending on $V(p,q)$:
\begin{equation}
\varphi_q(0) \;=\;
\begin{cases}
0 & \text{if $V(p,q)=1$,}\\
1 & \text{if $V(p,q)=0$.}
\end{cases}
\end{equation}
To make this rigorous, define a total computable function $f:\mathbb{N}\to\mathbb{N}$ as follows:
given $e\in\mathbb{N}$, let $f(e)$ be an index of the program that on input $0$ computes $V(p,e)$ and outputs $0$ if $V(p,e)=1$, and outputs $1$ otherwise.
By the $s$-$m$-$n$ theorem, such an index can be computed effectively from $e$: the map $f$ is a purely syntactic program transformation, and it is total computable irrespective of the behavior of $V$.
(The totality of $V$ enters separately: it guarantees that the program with index $f(e)$ in fact halts on input $0$.)
By Kleene's Recursion Theorem, there exists an index $q$ such that
\begin{equation}
\varphi_q \;=\; \varphi_{f(q)}.
\end{equation}
In particular, on input $0$, program $q$ behaves exactly as specified above:
it computes $V(p,q)$ and outputs $0$ if $V(p,q)=1$, else outputs $1$.
There are now two cases:

\smallskip
\noindent\emph{Case A: $V(p,q)=1$.}
Then by the definition of $q$, $\varphi_q(0)$ halts and outputs $0$, hence $J(q)=0$.
Since $J(p)=0$, we have that $J(q)>J(p)$ is false.
This contradicts \textbf{soundness}, because $V(p,q)=1$ but $q$ is not a strict improvement over $p$.

\smallskip
\noindent\emph{Case B: $V(p,q)=0$.}
Then $\varphi_q(0)$ halts and outputs $1$, hence $J(q)=1$.
Since $J(p)=0$, we have that $J(q)>J(p)$ is true.
This contradicts \textbf{completeness}, because $q$ is a strict improvement over $p$ but $V(p,q)\neq 1$.

\smallskip
In both cases we reach a contradiction, so no such total computable certifier $V$ can exist.
\end{proof}

\paragraph{Remark (Generality of the semantic objective).}
Theorem~1 uses a particularly simple semantic score $J$ (halting-and-output-$1$ on input $0$) to make the diagonal construction transparent.
The construction, however, uses only two properties of the score, and it is worth stating them exactly.
Let $\mathcal{S}$ carry a strict order $>$ and let $J:\mathbb{N}\to\mathcal{S}$ satisfy:
\emph{(i) extensionality}: $J(p)$ depends only on the partial function $\varphi_p$, not on the program text (so we may write $J$ on partial computable functions); and
\emph{(ii) attainable comparability}: there exist partial computable functions $\psi_{\mathrm{lo}},\psi_{\mathrm{hi}}$ with $J(\psi_{\mathrm{lo}})<J(\psi_{\mathrm{hi}})$.
Then no total computable certifier is sound and complete for strict improvement with respect to $J$:
fix a base index $p$ with $\varphi_p=\psi_{\mathrm{lo}}$, and let $f(e)$ be an index of the program that first computes $V(p,e)$ and then simulates a fixed program for $\psi_{\mathrm{lo}}$ if $V(p,e)=1$ and one for $\psi_{\mathrm{hi}}$ otherwise; the argument of Theorem~1 goes through verbatim.
Each hypothesis enters at exactly one point: extensionality makes the flip meaningful (the two branches realize the two $J$-values);
attainable comparability is the non-triviality requirement (note that non-constancy alone does not suffice if $>$ is a partial order --- the two attained values must be comparable --- and that for a constant $J$ the certifier $V\equiv 0$ is trivially sound and complete);
the order on $\mathcal{S}$ enters only through the single comparison $J(\psi_{\mathrm{hi}})>J(\psi_{\mathrm{lo}})$;
and totality of $V$ is used exactly once, to ensure the constructed program halts.
This is the Rice-style phenomenon \citep{rice1953classes} in relational form; the specificity of $J$ in Theorem~1 is for concreteness only.

\paragraph{Remark (Relation to Rice's theorem).}
For the certifier alone, Theorem~1 is close to a restatement of Rice's theorem.
Once the base agent $p$ is fixed, strict improvement is a nontrivial extensional property of the candidate, so no total computable procedure can decide it.
The previous remark makes this precise.
The paper adds three things the classical statement does not give.
First, the proof is constructive.
Rice's theorem only states that no decider exists.
The recursion-theorem argument builds a concrete candidate that is good if rejected and bad if accepted.
Such a candidate uses the gate as an oracle.
This is the formal counterpart of evaluator hacking, and the same construction carries over to search (Theorem~2) and to multi-agent pipelines (Appendix~\ref{app:multiagent}).
Second, the search task of Theorem~2 looks weaker than a decision task.
A searcher does not have to classify every candidate.
It only has to return one improving proposal when the enumerator contains one, and it may even read the enumerator's code.
Rice's theorem says nothing about finding witnesses with access to the generator's code.
Theorem~2 shows that the weaker task is impossible as well, because the proposal mechanism itself can host the diagonal.
Third, undecidability results do not say how to restore guarantees.
Theorem~3 and the remark after it do.
Guarantees return once the gate commits to budgets it can compute, and they hold relative to the bounded score $J_T$.

\begin{figure}[t]
\centering
\begin{tikzpicture}[
  box/.style={draw, rounded corners, align=center, inner sep=6pt},
  arr/.style={-Latex, thick}
]
\node[box] (S) {Searcher $S$\\inputs: $(p,e)$};
\node[box, right=2.6cm of S] (E) {Enumerator $E_e$\\proposes $q=E_e(p,0)$};
\node[box, below=1.6cm of E] (q) {Candidate $q$\\uses $S(p,e)$};

\draw[arr] (S) -- node[above]{queries} (E);
\draw[arr] (E) -- node[right]{outputs $q$} (q);
\draw[arr] (q) -- node[below left]{depends on} (S);
\end{tikzpicture}
\caption{Self-referential ``diagonal'' structure in Theorem~2: the proposal mechanism outputs a candidate whose behavior depends on the searcher's decision.}
\label{fig:diagonal}
\end{figure}
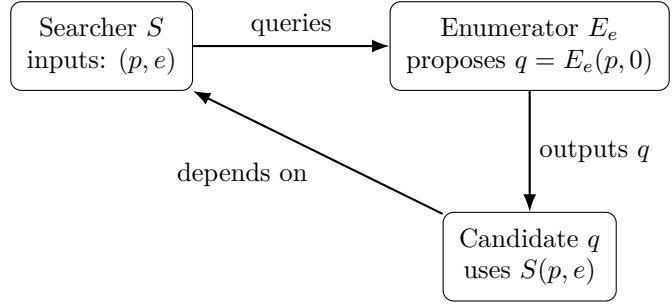

\section{Searching for Improvements}

Theorem~1 rules out a total computable certifier that accepts \emph{exactly} the strictly improving updates.
A natural response is to weaken the requirement: instead of certifying every candidate update, we only ask for
a total computable \emph{searcher} that returns \emph{some} improvement whenever the proposal mechanism contains one.
We show that even this goal is impossible in full generality.

\begin{definition}[Proposal enumerator]
A \emph{proposal enumerator} is a total computable function
\begin{equation}
E:\mathbb{N}\times\mathbb{N}\to\mathbb{N},
\end{equation}
where $E(p,i)$ denotes the $i$-th proposed modification for base agent $p$.
\end{definition}

\begin{definition}[Total searcher; proposal-respecting soundness and weak completeness]
A \emph{total searcher} is a total computable function
\[
S:\mathbb{N}\times\mathbb{N}\to (\mathbb{N}\cup\{\bot\}),
\]
where $\bot$ denotes ``no improvement found'' and the second argument is an index $e$ encoding a proposal enumerator $E_e$.
We restrict attention to \emph{proposal-respecting} searchers: if $S(p,e)=q\neq\bot$, then
$q = E_e(p,i)$ for some $i\in\mathbb{N}$.

We require:
\begin{itemize}
    \item \emph{Soundness:} If $S(p,e)=q\neq\bot$, then $J(q)>J(p)$.
    \item \emph{Weak completeness:} If $\exists i$ such that $J(E_e(p,i))>J(p)$, then $S(p,e)\neq\bot$.
\end{itemize}
\end{definition}

\begin{theorem}[No sound and weakly complete proposal-respecting total searcher]
Fix the score functional $J$ from Theorem~1.
There is no proposal-respecting total computable searcher $S$ that is sound and weakly complete for all proposal enumerators.
Equivalently, for every proposal-respecting total computable $S$, there exist a base agent $p$ and an enumerator index $e$
such that $S$ violates soundness or weak completeness on $(p,e)$.
\end{theorem}

\begin{proof}
Assume for contradiction that such a total computable searcher $S$ exists.
Fix a baseline program $p$ with $J(p)=0$ (e.g., $\varphi_p(0)$ halts and outputs $0$).
We will construct an enumerator $E$ (given by some index $e^\star$) that proposes a distinguished candidate
$q = E(p,0)$ designed to be adversarial to $S(p,e^\star)$.
For each index $e$, let $q_e$ be the program that on input $0$:
\begin{enumerate}
    \item computes $r \leftarrow S(p,e)$,
    \item computes $c \leftarrow E_e(p,0)$,
    \item outputs $0$ if $r=c$, and outputs $1$ otherwise.
\end{enumerate}
The map $e \mapsto q_e$ is total computable: by the $s$-$m$-$n$ theorem it is a purely syntactic transformation that hard-wires $e$ into a fixed program template, irrespective of the behavior of $S$ or $E_e$.
(Totality of $S$, and totality of the enumerator at the fixed point constructed below, matter only for the \emph{runtime} behavior of $q_e$; for an arbitrary index $e$, the computation of $E_e(p,0)$ inside $q_e$ may diverge, which is harmless for the computability of the map.)
Define now a total computable transformation $g$ mapping an index $e$ to an index $g(e)$ of an enumerator $E_{g(e)}$
satisfying:
\begin{equation}
E_{g(e)}(p,0) = q_e,
\qquad
E_{g(e)}(p,i) = p \ \text{ for all } i>0,
\end{equation}
and for all other base agents $p'\neq p$, set $E_{g(e)}(p',i)=p$ (any fixed dummy index suffices).
Such a transformation $g$ exists and is total computable by standard effective programming operations
(e.g., via the $s$-$m$-$n$ theorem); note that $E_{g(e)}$ is total for \emph{every} $e$, since producing the index $q_e$ from $e$ is itself a syntactic operation.

By Kleene's Recursion Theorem, there exists an index $e^\star$ such that
\begin{equation}
E_{e^\star} = E_{g(e^\star)}.
\end{equation}
Let $E := E_{e^\star}$ and let $q := E(p,0)$.
By construction and the fixed-point property, we have $q = q_{e^\star}$.
Similar to the proof of Theorem~1, consider now the output of the searcher $S(p,e^\star)$:

\smallskip
\noindent\emph{Case A: $S(p,e^\star)=q$.}
Then, by the definition of $q=q_{e^\star}$, the program $q$ outputs $0$ on input $0$, hence $J(q)=0$.
Since $J(p)=0$, we have that $J(q)>J(p)$ is false, contradicting \textbf{soundness}.

\smallskip
\noindent\emph{Case B: $S(p,e^\star)\neq q$.}
Then $q$ outputs $1$ on input $0$, hence $J(q)=1$ and $J(q)>J(p)$ holds.
Moreover, $q$ appears in the proposal list as $E(p,0)$, so there exists $i$ (namely $i=0$) with
$J(E(p,i))>J(p)$. By weak completeness, $S(p,e^\star)\neq\bot$.

Since $E(p,i)=p$ for all $i>0$, every proposal other than $q$ has score $J(p)=0$ and is \emph{not} a strict improvement.
Thus, under the assumption $S(p,e^\star)\neq q$, the (non-$\bot$) output of $S$ must be a non-improving candidate, which violates \textbf{soundness}.

\smallskip
In both cases, we reach a contradiction. Therefore, no such total computable searcher $S$ exists.
\end{proof}

\section{Randomized, Hidden, and Stateful Gates}
\label{sec:variants}

Real gates rarely look like the deterministic certifier of Theorem~1.
A reward model may use randomness.
Its weights may be hidden from the proposer.
It may keep state across queries.
This section explains why these variations do not escape the impossibility.

\paragraph{Randomized gates.}
Suppose the gate may flip fair coins.
We model it as a probabilistic machine $V$ that on input $(p,q)$ halts on every coin sequence and outputs a bit.
We do not ask for exact decisions, only for bounded error.

\begin{proposition}[No randomized certifier with bounded error]
\label{prop:randomized}
Let $J$ be the score functional from Theorem~1.
There is no probabilistic machine $V$ that halts on every input and every coin sequence and satisfies, for all $p,q$:
\begin{enumerate}
  \item if $J(q)>J(p)$, then $\Pr[V(p,q)=1]\ge 2/3$,
  \item if $J(q)>J(p)$ does not hold, then $\Pr[V(p,q)=1]\le 1/3$.
\end{enumerate}
\end{proposition}

\begin{proof}
Suppose such a $V$ exists.
Fix an input $(p,q)$ and consider the binary tree of coin sequences explored by $V$.
Every path halts, so by K\"onig's lemma the tree is finite.
A deterministic machine can therefore simulate all paths and compute the exact acceptance probability $\alpha(p,q)$, a rational number.
Define $V'(p,q)=1$ if $\alpha(p,q)\ge 1/2$ and $V'(p,q)=0$ otherwise.
Then $V'$ is total computable, and the error bounds make it sound and complete for strict improvement.
This contradicts Theorem~1.
\end{proof}

The assumption that every coin sequence halts can be weakened.
\citet{deleeuw1956computability} show that machines that halt with probability one and err with probability bounded below $1/2$ decide only decidable sets, so the same conclusion holds.
The same simulation also derandomizes any randomized searcher that returns, for every input, one fixed value with probability at least $2/3$.
The deterministic version returns that value, so Theorem~2 rules it out as well.
The lesson is simple.
Randomness can hide the gate's decision from the proposer, but it cannot enlarge the class of relations the gate decides.

\paragraph{Hidden gates.}
Theorems~1 and~2 quantify over gates.
For every gate there exists a candidate on which it fails, and the proof builds this candidate from the code of the gate.
Hiding the gate changes who can carry out this construction.
A proposer without access to the gate's code cannot build the diagonal candidate.
It does not change whether the failure exists.
The guarantee ``sound and complete for all inputs'' is still false, because the bad input pair exists whether or not anyone can find it.
Hiddenness is therefore a defense against a specific attacker, not a repair of the guarantee.
It also degrades in practice: query access, accept and reject outcomes, and leaked transcripts all let a proposer learn a predictive model of the gate (Appendix~\ref{app:multiagent}).

\paragraph{Stateful gates.}
A gate that keeps state across queries is a machine with an extra state argument.
Fix any reachable state $s$ and consider the total computable certifier $V_s(p,q)=V(p,q,s)$.
Theorem~1 applies to $V_s$.
So for every state the gate can be in, there is a candidate on which it fails in that state.
State reshuffles which candidate defeats the gate, but some candidate always does.

\paragraph{Human oversight.}
A gate that consults a human is not a total computable procedure, so our theorems do not constrain it.
This is a design lesson rather than a loophole.
The impossibility concerns fully automated gates, and it bounds what automation alone can guarantee.
Grounding decisions outside the loop is exactly what the theorems leave open.

\section{Decidability under Bounded Evaluation and Finite Proposals}
\label{sec:bounded}

The negative results above rely on unbounded program behavior and universally quantified proposal mechanisms.
We now formalize a standard restriction used in practice: bounded evaluation and finite search.

\begin{definition}[Bounded-horizon score]
Define a time bound $T\in\mathbb{N}$ and bounded score $J_T:\mathbb{N}\to\{0,1\}$ by
\begin{equation}
J_T(p)=
\begin{cases}
1 & \text{if $\varphi_p(0)$ halts within $T$ steps and outputs $1$,}\\
0 & \text{otherwise.}
\end{cases}
\end{equation}
\end{definition}

\begin{theorem}[Sound and weakly complete total search under bounded horizon and finite proposals]
Fix an evaluation bound $T\in\mathbb{N}$, a proposal budget $B\in\mathbb{N}$, and a generation bound $T_E\in\mathbb{N}$, and let $J_T$ be as above.
There exists a total computable searcher $S_{T,B,T_E}$ such that for \emph{all} inputs $p,e\in\mathbb{N}$:
\begin{enumerate}
  \item \textbf{Soundness:} if $S_{T,B,T_E}(p,e)=q\neq\bot$, then $q=E_e(p,i)$ for some $i<B$ whose computation halts within $T_E$ steps, and $J_T(q)>J_T(p)$.
  \item \textbf{Weak completeness within budget:} if the computations $E_e(p,i)$ halt within $T_E$ steps for all $i<B$, and there exists $i<B$ with $J_T(E_e(p,i))>J_T(p)$, then $S_{T,B,T_E}(p,e)\neq\bot$.
\end{enumerate}
\end{theorem}

\begin{proof}
Define $S_{T,B,T_E}(p,e)$ as follows.
First compute $J_T(p)$ by simulating $\varphi_p(0)$ for at most $T$ steps.
Then, for each $i=0,1,\dots,B-1$: simulate the computation of $E_e(p,i)$ for at most $T_E$ steps; if it does not halt within this budget, skip this $i$; otherwise, let $q_i$ be its output and compute $J_T(q_i)$ by simulating $\varphi_{q_i}(0)$ for at most $T$ steps.
Return the first $q_i$ with $J_T(q_i)>J_T(p)$; if no $i<B$ yields one, return $\bot$.
Every operation is a step-bounded simulation and the loop is bounded by $B$, so the procedure halts on \emph{every} input (including indices $e$ that do not encode total enumerators) and is therefore total computable.
Soundness and weak completeness within budget follow immediately from the construction.
\end{proof}

\paragraph{Remark (The role of the bounds).}
Two comments.
First, the generation bound $T_E$ is what makes $S_{T,B,T_E}$ total on its \emph{full} domain: without it, merely evaluating $E_e(p,i)$ on an arbitrary index $e$ could diverge, so a searcher required to be total on all inputs must guard proposal generation exactly as it guards candidate evaluation.
Second, nothing in Theorem~3 requires the bounds to be constants: $T$, $B$, and $T_E$ may be replaced by arbitrary total computable functions $T(p,e)$, $B(p,e)$, $T_E(p,e)$ --- the searcher first computes the bound values and then runs the same exhaustive procedure, so not even time-constructibility in the complexity-theoretic sense is needed.
The boundary drawn by Theorems~1--3 is therefore not ``constant versus growing budgets'' but ``budgets the gate can compute versus unbounded semantic behavior'': the diagonal argument returns exactly when the score depends on behavior beyond every bound the gate commits to in advance.

\section{Repeated Updates and Proxy Drift}
\label{sec:drift}

Theorems~1 to~3 concern a single update step.
Self-improvement is a loop.
The current agent is replaced, the new agent proposes again, and the process repeats.
This section shows that the guarantees of Theorem~3 do not compose across rounds.
Each accepted step can improve the bounded score while the true score falls.

To state this, we extend the scores of Theorems~1 and~3 from one input to several.
Fix $n\in\mathbb{N}$ and define
\begin{equation}
J^{(n)}(p)=\bigl|\{x<n:\varphi_p(x) \text{ halts and outputs } 1\}\bigr|,
\end{equation}
\begin{equation}
J^{(n)}_T(p)=\bigl|\{x<n:\varphi_p(x) \text{ halts within } T \text{ steps and outputs } 1\}\bigr|.
\end{equation}
Think of $n$ tasks and a benchmark that scores each task under a timeout.
The bounded searcher of Theorem~3 carries over, since $J^{(n)}_T$ is total computable by $n$ step-bounded simulations.
Exhaustive search is therefore again sound and weakly complete for $J^{(n)}_T$.

\begin{proposition}[Accepted updates can lower the true score]
\label{prop:drift}
Let $n=2k$ for some $k\ge 1$, and let $T$ be large enough that a program can read an input $x<n$ and output $1$ within $T$ steps.
Then there are programs $p_0,p_1,\dots,p_k$ such that for every round $i<k$:
\begin{enumerate}
  \item $J^{(n)}_T(p_{i+1}) > J^{(n)}_T(p_i)$, so a gate that is sound and complete for the bounded score accepts the update, and
  \item $J^{(n)}(p_{i+1}) < J^{(n)}(p_i)$, so the true score falls.
\end{enumerate}
Over the whole run, the bounded score rises from $0$ to $k$ while the true score falls from $2k$ to $k$.
\end{proposition}

\begin{proof}
Define $p_i$ as the program that on input $x<n$ acts as follows.
If $x<i$, it outputs $1$ immediately.
If $i\le x<n-i$, it idles for $T+1$ steps and then outputs $1$.
If $x\ge n-i$, it outputs $0$.
The bound $T$ and the thresholds are fixed constants wired into the program, so each $p_i$ exists.
The first case contributes $i$ fast successes, and the second case contributes $n-2i$ slow successes.
Hence $J^{(n)}_T(p_i)=i$ and $J^{(n)}(p_i)=i+(n-2i)=n-i$.
Both claims follow by comparing consecutive rounds.
\end{proof}

The mechanism is plain.
Each update trades slow correct behavior for fast correct behavior on fewer tasks.
The gate cannot see the loss, because the loss lives beyond its time budget.
No self-reference and no adversary is involved.
The chain sits entirely inside the decidable regime of Theorem~3, so boundedness blocks the diagonal of Theorems~1 and~2 but not drift.
The two failure modes are different, and fixing one does not fix the other.

Drift is visible to a larger budget.
As soon as some later evaluation uses a bound above $T$ plus the idle delay, the slow successes reappear and the loss shows.
This gives a simple design rule for improvement loops: hold the per-round budget fixed for speed, and re-evaluate with a larger budget from time to time.
How fast drift can grow when the proposer adapts to the re-evaluation schedule is open.

\section{Discussion}
\label{sec:discussion}

\paragraph{Why the diagonal breaks under bounded evaluation.}
In the proofs of Theorems~1--2, the adversarial candidate $q$ runs the gate ($V$ or $S$) as a subroutine.
In the bounded regime of Theorem~3 this no longer works.
The searcher $S_{T,B,T_E}$ simulates each candidate for up to $T$ steps across $B$ proposals, so its own running time exceeds $T$.
A candidate that runs $S_{T,B,T_E}$ inside itself therefore cannot halt within $T$ steps.
Its bounded score is $J_T(q)=0$, and the self-referential flip no longer yields a strict improvement.
This is how bounded evaluation blocks the diagonal.

\paragraph{What the theorems do and do not claim.}
Theorems~1--2 are worst-case results.
They rule out universal procedures that always halt and are sound and complete (or weakly complete) for all sufficiently expressive agents and proposal mechanisms.
They do not imply that self-improvement cannot occur.
They also do not imply that useful gates cannot be built for restricted settings.
What they show is a gap between semantic correctness, as measured by $J$, and any terminating gate that must work uniformly over all candidates and generators.

\paragraph{Where undecidability enters.}
The diagonal constructions in Theorems~1--2 use two modeling freedoms.
Agents may run unboundedly, and candidates may depend on the evaluation procedure through self-reference.
Both freedoms are natural for tool-using agents that execute code and adapt their patches.
Theorem~2 even hands the searcher the code of the proposal enumerator.
A total method that is sound and weakly complete for all generators still does not exist.

\paragraph{One-sided guarantees and intermediate regimes.}
Only the joint requirement fails.
Each guarantee alone is trivial.
The gate that rejects every update is sound, and the gate that accepts every update is complete.
A more useful one-sided design is proof-carrying updates.
Here the gate accepts $q$ only when the proposal comes with a certificate, checked by a fixed sound proof system, that $J(q)>J(p)$.
The G\"odel machine of \citet{schmidhuber2003goedel} is the canonical design of this kind: it rewrites its own code only after an internal searcher finds a proof that the rewrite raises expected utility.
Such a gate is total and sound by construction, but it cannot be complete.
For the score of Theorem~1, the set of true improvement pairs $\{(p,q): J(q)>J(p)\}$ is not computably enumerable, since witnessing $J(p)=0$ is a $\Pi_1$ fact.
The set of pairs any fixed proof-checking gate accepts is computably enumerable.
Such a gate therefore misses infinitely many genuine improvements.
This gives a precise account of the conservatism of proof-gated self-improvement.
Between the unbounded setting of Theorems~1--2 and the finite regime of Theorem~3 lie further restrictions.
Section~\ref{sec:variants} treats three of them: randomized, hidden, and stateful gates.
Other examples are proposal mechanisms that cannot condition on the gate (Appendix~\ref{app:multiagent}), update languages whose effect on the score is decidable, and gates with more compute than the proposer.
Mapping which restriction restores which guarantee is a concrete open problem.

\paragraph{Decidable regimes.}
Section~\ref{sec:bounded} formalizes a restriction that practice already uses, namely bounded evaluation and finite search.
Under it, sound and weakly complete search exists (Theorem~3).
The impossibility therefore comes from the unbounded and universal setting alone.
Explicit design choices such as timeouts and finite candidate budgets remove it.

\paragraph{Implications for evaluation pipelines.}
In ML terms, $V$ and $S$ stand for reward models, test suites, and automated code review.
Theorems~1--2 show that no single terminating mechanism can bridge proxy evaluation and semantic improvement once behavior is unbounded and proposals can self-refer.
Reliable pipelines therefore need combinations of bounded evaluation, restricted update languages, staged deployment and sandboxing, and one-sided checks that trade completeness for safety.

\paragraph{Implications for automated research pipelines.}
Recent work builds end-to-end pipelines in which a language-model ``scientist'' generates ideas and code, runs experiments, and a second model or a scripted rubric reviews the outcome (e.g., \emph{The AI Scientist} \citep{lu2024ai, yamada2025ai}).
Such a pipeline matches our model.
The idea generator and the experiment runner form a proposal mechanism, and the reviewer acts as a computable gate.

Our results do not claim that any particular system is broken.
They state a worst-case limit that such systems meet as their autonomy grows.
Suppose the proposal mechanism can condition on the gate, for example by learning to predict the reviewer's preferences.
Then no terminating gate can guarantee both that it never accepts a non-improvement and that it never rejects or misses an available one, uniformly over all proposals.
This is the regime where benchmark overfitting, reward hacking, and self-deception enter.
The system optimizes what the reviewer measures, not what the designer intends.

At the same time, automated research loops run with bounded compute and finite candidate budgets, so they sit in the decidable regime of Theorem~3.
Exhaustive evaluation within a fixed budget then gives reliable guarantees with respect to the bounded proxy.
The remaining risk is the gap between the bounded proxy and the intended objective.
Section~\ref{sec:drift} shows that this risk is real even without an adversary, since accepted updates can raise the benchmark score while the true score falls.
This motivates design choices already common in practice, such as timeouts, finite proposal budgets, restricted update languages, sandboxing, and occasional human grounding \citep{jimenez2023swe}.

\paragraph{Interactive and environment-coupled scores.}
A natural worry is that real agents are evaluated through interaction with an environment, while $J$ in Theorem~1 looks only at $\varphi_p(0)$.
For the negative results, this restriction is harmless.
The score $J$ may be any functional that depends only on the partial function $\varphi_p$.
Examples are the expected reward of $p$ in a stochastic environment over a fixed horizon, the success rate on a benchmark distribution, or any scalar summary of trajectories obtained by reading $\varphi_p$ as a policy over observation histories.
As long as $J$ is extensional and two comparable values are attainable, the diagonal of Theorem~1 applies, and Theorem~2 extends to interactive search in the same way.
A certifier that fails for a static program also fails when evaluation must simulate an adaptive environment on top.

The positive result needs more care.
A finite horizon alone does not make expected reward computable.
Theorem~3 extends to interactive evaluation only under explicit assumptions: finitely many actions and observations, computable transition probabilities with rational values, computable rational rewards, and a cap on the agent's compute per step.
Under these assumptions the evaluation tree is finite and every quantity in it is rational, so the interactive score is total computable and the searcher of Theorem~3 carries over (Proposition~\ref{prop:interactive} in Appendix~\ref{app:interactive}).
Without them, exact comparison of scores can itself be undecidable, and only approximate or one-sided guarantees remain.

\section{Related Work}

Our results sit within classical limits in computability theory \citep{cooper2017computability}.
The impossibility of a universal improvement certifier is aligned with the broader principle that non-trivial semantic properties of programs are undecidable, as captured by Rice's theorem \citep{rice1953classes}.
While Rice's theorem is often stated for properties of single programs, our setting concerns a relational property between a base agent and a proposed modification, and we use explicit fixed-point constructions tailored to this verification-gated ``self-improvement'' interpretation.
We emphasize that the technical tools are classical; the novelty is the ML-facing formulation (i.e., modeling automated self-improvement as a gated update loop with explicit reliability and progress guarantees) and the resulting boundary between an unbounded regime where such guarantees cannot hold uniformly and a bounded regime where they can be recovered by design.

The self-reference used in our constructions is enabled by Kleene's Recursion Theorem \citep{kleene1952introduction}, a standard tool for building programs whose behavior depends on their own descriptions.
In AI safety and agent foundations, related obstacles arise when agents attempt to reason about, verify, or safely delegate to more capable successors.
Work on self-referential agents, including the L\"obian obstacle, tiling agents, and Vingean reflection, studies conditions under which an agent can reason about successor systems without collapsing into inconsistency or conservatism \citep{soares2015toward,yudkowsky2013tiling}.
Our contribution is to translate these computability-theoretic phenomena into the concrete language of modern ML evaluation pipelines (reward models, test suites, and automated reviewers), and to show that replacing certification with search does not avoid the limitation when proposal mechanisms can adapt to the evaluator.

Two lines of prior work address self-improvement directly.
\citet{schmidhuber2003goedel} proposes the G\"odel machine, a self-referential system that rewrites any part of its own code once an internal proof searcher shows that the rewrite raises expected utility.
In our terminology this is a proof-carrying gate.
It is sound by construction, and the Discussion (Section~\ref{sec:discussion}) locates its unavoidable cost: the set of provable improvements is computably enumerable while the set of true improvements is not, so every such gate misses infinitely many true improvements.
\citet{everitt2016self} formalize agents that can modify their own policies and utility functions and study when a rational agent will preserve or subvert its goals.
Their question is decision-theoretic and takes the evaluation of outcomes as given.
Our question is computational: whether any terminating procedure can perform that evaluation at all.
The two are complementary, and the drift result of Section~\ref{sec:drift} sits between them, since it shows the true score falling through the gap between a proxy and the objective, without any self-modification of goals.

A complementary line of work studies the empirical and conceptual phenomenon whereby optimizing a proxy objective diverges from the intended goal.
\citet{amodei2016concrete} identify reward hacking as a core problem in AI safety, observing that agents can exploit misspecified reward signals in ways unintended by the designer.
\citet{manheim2019categorizing} formalize several variants of Goodhart's law for machine learning, distinguishing regressional, extremal, causal, and adversarial failure modes by which a proxy metric ceases to track the true objective under optimization pressure.
\citet{skalse2022defining} provide a formal framework for reward hacking in reinforcement learning, defining conditions under which a proxy reward function can be exploited by a sufficiently capable policy, and showing that such exploitation is generic rather than pathological.
More recently, \citet{pan2022effects} empirically demonstrate that reward model overoptimization degrades true performance as optimization pressure increases, even when the proxy initially correlates well with the ground truth.

Our results complement this literature from a computability-theoretic angle.
The reward hacking literature typically studies \emph{statistical} proxy-objective gaps: the proxy is computable but misspecified, and failure arises from distributional shift or optimization pressure.
Our impossibility results show that even if the proxy were \emph{perfectly specified}, i.e., the certifier or searcher attempts to track the true semantic score $J$ exactly, no total computable gate can be simultaneously sound and complete when candidates can self-referentially depend on the gate.
In other words, the gap we identify is not statistical but \emph{structural}: self-reference introduces a failure mode that cannot be closed by better proxy design alone, only by restricting the computational regime (bounded evaluation, finite proposals) as formalized in Theorem~3.

\section{Conclusion \& Future Work}

We studied a minimal formal model of verification-gated self-improvement in which agents are programs, proposed updates are generated by a computable mechanism, and acceptance decisions are made by a terminating gate.
The central finding is a sharp limitation in the unbounded, fully general setting: no terminating procedure can simultaneously guarantee that it never accepts a non-improvement and that it never rejects or misses a genuine improvement when proposals may be self-referential.
This failure is structural rather than statistical.
It follows from the interaction between unbounded program behavior and the ability of a proposal mechanism to adapt to the evaluator.
It also survives randomization with bounded error and hiding the gate from the proposer (Section~\ref{sec:variants}).

At the same time, the negative results identify exactly which restrictions restore decidability.
When evaluation is explicitly bounded and the proposal budget is finite, exhaustive checking recovers a terminating procedure with the desired guarantees relative to the bounded score being measured. The overall results motivate a simple design principle for autonomous improvement loops: make the limits explicit.
Timeouts, finite candidate budgets, restricted update languages, sandboxing, and occasional external or human grounding are not optional conveniences.
They are the conditions under which correctness guarantees become attainable.

Section~\ref{sec:drift} takes a first step beyond the single update step and shows that accepted updates can raise the bounded score in every round while the true score falls.
An important direction for future work is to characterize how fast this drift can grow when the proposer adapts to the evaluation schedule, and which additional constraints can control it while retaining useful progress.

\clearpage
\newpage

\bibliography{main}
\bibliographystyle{tmlr}

\appendix
\section{Multi-Agents}
\label{app:multiagent}

This appendix explains why the impossibility results extend to multi-agent pipelines.
The main point is that (i) an interacting collection of agents can be encoded as a single composite program, and
(ii) even if we keep the agents conceptually separate, self-reference can be realized through message passing
whenever the proposer can condition on the gate's behavior (directly or indirectly).

\subsection{A minimal multi-agent model}
We model a multi-agent self-improvement pipeline as a finite tuple of programs together with a fixed interaction protocol.
Let
\[
\mathcal{A} = (P, R, O, \Pi)
\]
denote a system consisting of a \emph{proposer} $P$, a \emph{reviewer} (or gate component) $R$, an \emph{orchestrator} $O$ (which routes messages, runs tests, and applies updates), and a protocol $\Pi$ that specifies how these components interact over discrete time steps.
On each round, the orchestrator provides inputs (task instances, tests, prompts, or environment states) to the components, collects their outputs, and produces an observable transcript.

We consider updates that replace one component while holding the others fixed (the most common case in practice), e.g.,
replacing $P$ with a candidate proposer $P'$.
Write $\mathcal{A}[P \leftarrow P']$ for the system where only the proposer is swapped.
A \emph{gate} is a terminating procedure that, given the current system and a candidate update, returns a binary accept/reject decision.
In the simplest version:
\[
G(\mathcal{A}, P') \in \{0,1\},
\]
where $G(\mathcal{A}, P')=1$ means ``accept replacing $P$ with $P'$.''

\subsection{Encoding a multi-agent system as a single program}
Because each component is a program and the protocol $\Pi$ is fixed, the joint behavior of $\mathcal{A}$ can be implemented by a single program that simulates the interaction internally.
Concretely, define a compiler (an effective transformation) that maps any tuple $(P,R,O,\Pi)$ to a single program index $\langle P,R,O,\Pi\rangle$ which:
(i) simulates the protocol $\Pi$ step-by-step,
(ii) runs the component programs on the appropriate messages, and
(iii) outputs the same external behavior (actions, transcript, or final decision) as the original multi-agent system.

This is standard: multi-tape machines, communicating processes, and modular code can be compiled into a single Turing machine with only polynomial overhead.
Therefore, any semantic score assigned to the multi-agent system (e.g., success probability, reward, correctness of a scientific claim) can be treated as a semantic score of the compiled single program.
In particular, if we define a semantic score functional over composite systems, we can equivalently define a semantic score over their single-program encodings.

\subsection{Why the diagonal channel persists in multi-agent form}
The negative results rely on a structural ingredient: the candidate update can be chosen so that its behavior depends on the gate's output.
In a multi-agent pipeline, such dependence can be realized through interaction in several natural ways:

\begin{itemize}
    \item \textbf{Direct query access.} The proposer can explicitly query the reviewer (or a proxy of the gate) during generation (e.g., ``would this be accepted?'').
    \item \textbf{Outcome feedback.} Even without explicit queries, the proposer can adapt using accept/reject outcomes from previous rounds.
    \item \textbf{Transcript leakage.} If reviewer comments or rubric scores are revealed, the proposer can learn a predictive model of acceptance and condition on it.
\end{itemize}

Any of these channels allows the proposer to implement a ``flip on accept'' strategy: construct a candidate $P'$ whose behavior under evaluation depends on whether $G(\mathcal{A},P')$ would accept it.
Intuitively, the proposer uses the gate as an oracle (directly or indirectly), creating a self-referential loop routed through other agents.

\subsection{A proof via reduction to the single-program setting}
\label{app:multiagent:a4}

We make the reduction from a fixed multi-agent pipeline to the single-program setting explicit and then apply
Kleene's Recursion Theorem.
Throughout, gates and scores operate on \emph{program indices} (finite syntactic objects), never on partial functions directly.

\paragraph{Setup.}
Fix a multi-agent system
\(
\mathcal{A}=(P,R,O,\Pi)
\)
and consider updates that replace only the proposer component.
For a proposer index \(e\in\mathbb{N}\), write \(\mathcal{A}[P\leftarrow e]\) for the (finite encoding of the) system in which the proposer component is the program with index \(e\).

\paragraph{Compilation to a single program.}
Because \(\Pi\) is fixed and each component is a program, there exists a total computable \emph{compiler}
\(
\mathsf{Comp}:\mathbb{N}\to\mathbb{N}
\)
such that for every proposer index \(e\),
\(
\mathsf{Comp}(e)
\)
is an index of a single program that simulates the composite system
\(
\mathcal{A}[P\leftarrow e]
\)
under protocol \(\Pi\) step by step.
Let \(J\) be an extensional semantic score on single-program indices, as in the main text, and define the induced score on proposer indices by
\begin{equation}
\label{eq:compiled-score}
\tilde J(e) \;:=\; J\!\left(\mathsf{Comp}(e)\right).
\end{equation}
Since the compiled program interacts with the proposer component only by running it on protocol messages, \(\varphi_{e}=\varphi_{e'}\) implies \(\varphi_{\mathsf{Comp}(e)}=\varphi_{\mathsf{Comp}(e')}\); combined with the extensionality of \(J\), the induced score \(\tilde J\) is itself extensional: it depends only on \(\varphi_e\).

\paragraph{Controllability assumption.}
Assume the following mild condition holds for this fixed pipeline:
there exist two proposer indices \(p_{\mathrm{bad}}, p_{\mathrm{good}}\in\mathbb{N}\) such that
\begin{equation}
\label{eq:controllability}
\tilde J(p_{\mathrm{bad}})=0,
\qquad
\tilde J(p_{\mathrm{good}})=1.
\end{equation}
(Equivalently, swapping only the proposer can induce at least two distinct semantic outcomes. This is satisfied by
the concrete \(J\) used in Theorem~1, and by essentially any task score for which ``obviously fail'' and
``obviously succeed'' proposer behaviors are achievable.)
Define the baseline system
\(
\mathcal{A}_0 := \mathcal{A}[P\leftarrow p_{\mathrm{bad}}]
\),
whose score is \(\tilde J(p_{\mathrm{bad}})=0\).

\paragraph{The gate.}
A gate is a total computable procedure \(G\) that receives the encoding of the current system and a candidate proposer \emph{index} and returns a bit.
Define the induced gate on proposer indices by
\begin{equation}
\label{eq:compiled-gate}
\tilde G(e) := G(\mathcal{A}_0, e).
\end{equation}
If \(G\) is total computable, then \(\tilde G\) is total computable.
Assume for contradiction that \(G\) is simultaneously
\emph{sound} and \emph{complete} for strict improvement over the baseline with respect to \(\tilde J\), i.e.,
\[
\tilde G(e)=1 \Rightarrow \tilde J(e) > \tilde J(p_{\mathrm{bad}})=0,
\qquad
\tilde J(e) > 0 \Rightarrow \tilde G(e)=1.
\]

\paragraph{Diagonal construction.}
Define a total computable function \(f:\mathbb{N}\to\mathbb{N}\) that maps an index \(e\) to an index
\(f(e)\) of a proposer program which, on any input:
\begin{enumerate}
\item computes \(b \leftarrow \tilde G(e)\) (this halts, since \(\tilde G\) is total),
\item if \(b=1\), simulates the program with index \(p_{\mathrm{bad}}\),
\item if \(b=0\), simulates the program with index \(p_{\mathrm{good}}\).
\end{enumerate}
By the \(s\)-\(m\)-\(n\) theorem, \(f\) is a purely syntactic transformation. It hard-wires \(e\) into a fixed program template. It is total computable irrespective of the behavior of \(\tilde G\); totality of \(\tilde G\) is used only so that the constructed program's first step halts.

\paragraph{Invoke Kleene's Recursion Theorem.}
By Kleene's Recursion Theorem, there exists an index \(e^\star\) such that
\begin{equation}
\label{eq:kleene-fixedpoint}
\varphi_{e^\star} = \varphi_{f(e^\star)}.
\end{equation}
Let \(e^\star\) be the corresponding proposer index.

\paragraph{Case analysis.}
Consider the value \(\tilde G(e^\star)=G(\mathcal{A}_0, e^\star)\).

\smallskip
\noindent\emph{Case A: \(\tilde G(e^\star)=1\).}
By the definition of \(f\) and the fixed-point property~\eqref{eq:kleene-fixedpoint}, \(\varphi_{e^\star}=\varphi_{p_{\mathrm{bad}}}\).
By extensionality of \(\tilde J\), it follows that \(\tilde J(e^\star)=\tilde J(p_{\mathrm{bad}})=0\).
Therefore \(e^\star\) is \emph{not} a strict improvement over the baseline, contradicting soundness.

\smallskip
\noindent\emph{Case B: \(\tilde G(e^\star)=0\).}
Again by the definition of \(f\) and~\eqref{eq:kleene-fixedpoint}, \(\varphi_{e^\star}=\varphi_{p_{\mathrm{good}}}\), hence by extensionality
\(\tilde J(e^\star)=\tilde J(p_{\mathrm{good}})=1>0=\tilde J(p_{\mathrm{bad}})\).
Thus \(e^\star\) \emph{is} a strict improvement over the baseline, contradicting completeness.

\smallskip
In both cases we obtain a contradiction. Hence no total computable gate \(G\) can be simultaneously sound and
complete for strict improvement in this multi-agent setting.
\qed

\subsection{When the extension does \emph{not} apply}
The diagonal channel requires that the candidate update can condition on the gate, either by querying it, observing outcomes, or learning from transcripts.
If the system is designed so that candidates are generated independently of the gate (no feedback, no shared transcript, no ability to model the reviewer), then the self-referential construction is blocked.
However, this restriction is strong and typically corresponds either to highly conservative pipelines or to bounded regimes where the gate's guarantees are explicitly limited to a fixed evaluation horizon and finite proposal budget.

\section{Interactive Evaluation with a Computable Environment}
\label{app:interactive}

The Discussion claims that Theorem~3 extends to interactive evaluation under explicit assumptions.
This appendix states those assumptions and proves the claim.

\paragraph{Setting.}
Fix a finite action set $A$ with a default action $a_0\in A$, a finite observation set $O$, a horizon $H\in\mathbb{N}$, and a per-step compute cap $T_{\mathrm{step}}\in\mathbb{N}$.
A history of length $t$ is a sequence $h=(a_1,o_1,\dots,a_t,o_t)$ of actions and observations, encoded as a natural number.
The environment is given by two total computable functions:
a transition function $\mu$ that maps a history and a next action to a probability distribution over $O$ with rational values, and
a reward function $r$ that maps a complete history of length $H$ to a rational number.
An agent is a program $p$.
At each step it receives the encoded current history, and we simulate $\varphi_p$ on it for at most $T_{\mathrm{step}}$ steps.
If the simulation halts with some $a\in A$, the agent plays $a$.
Otherwise the agent plays the default action $a_0$.
The \emph{interactive bounded score} $J^{\mathrm{int}}(p)$ is the expected value of $r$ over the distribution on length-$H$ histories induced by the agent and the environment.

\begin{proposition}[Computable interactive score]
\label{prop:interactive}
Under the assumptions above, $J^{\mathrm{int}}:\mathbb{N}\to\mathbb{Q}$ is total computable.
Consequently, the exhaustive searcher of Theorem~3, run with $J^{\mathrm{int}}$ in place of $J_T$, is total computable, sound, and weakly complete within budget with respect to $J^{\mathrm{int}}$.
\end{proposition}

\begin{proof}
Consider the tree of histories reachable in $H$ steps.
Its depth is $H$ and its branching is at most $|O|$ per step, so it has at most $|O|^H$ leaves.
At each node, the agent's action is determined by a simulation of at most $T_{\mathrm{step}}$ steps with a default on timeout, so it always terminates and is computable from $p$.
The probability of each edge is a rational value of $\mu$, and the value of each leaf is a rational value of $r$.
Therefore $J^{\mathrm{int}}(p)$ is a finite sum of products of rational numbers, each computable from $p$.
Comparisons between rationals are decidable, so the exhaustive procedure of Theorem~3 goes through unchanged.
\end{proof}

\paragraph{Each assumption does real work.}
The finite horizon keeps the tree finite in depth.
The finite action and observation sets keep the branching finite.
The per-step cap makes the agent's behavior at each node computable, since without it a single step could diverge.
The rational values matter for the comparisons.
If rewards or probabilities are arbitrary computable reals, the score can still be approximated to any precision, but deciding the strict inequality $J^{\mathrm{int}}(q)>J^{\mathrm{int}}(p)$ is no longer possible in general, because equality of computable reals is undecidable.
In that case a one-sided design remains available, for example accepting only when an approximation certifies a gap of at least a known margin.

\end{document}